\documentclass[conference]{IEEEtran}

\usepackage{cite}
\usepackage{amsmath, amssymb, amsfonts}
\usepackage{graphicx}
\usepackage{textcomp}
\usepackage{multirow}
\usepackage{array}
\usepackage{booktabs}
\usepackage{rotating}
\usepackage{hyperref}
\usepackage{fancyhdr}
\usepackage{lmodern}
\usepackage{adjustbox}
\usepackage{ragged2e}
\usepackage{balance}

\hypersetup{
    colorlinks=true,
    urlcolor=blue,
    citecolor=black,
    pdftitle={Overleaf Example},
    pdfpagemode=FullScreen,
}
\usepackage[utf8]{inputenc}  
\usepackage{textcomp}        

\DeclareUnicodeCharacter{22C6}{\textasteriskcentered}
\DeclareUnicodeCharacter{2212}{\ensuremath{-}}

\author{
\IEEEauthorblockN{
    Palak Handa\IEEEauthorrefmark{1}, 
    Manas Dhir\IEEEauthorrefmark{2}, 
    Amirreza Mahbod\IEEEauthorrefmark{1}, 
    Florian Schwarzhans\IEEEauthorrefmark{1}, \\
    Ramona Woitek\IEEEauthorrefmark{1}, 
    Nidhi Goel\IEEEauthorrefmark{3}, 
    Deepak Gunjan\IEEEauthorrefmark{4}, 
}
\vspace{5mm}
\IEEEauthorblockA{\IEEEauthorrefmark{1}Research Center for Medical Image Analysis and Artificial Intelligence, 
Department of Medicine, \\Danube Private University, Krems, Austria}
\IEEEauthorblockA{\IEEEauthorrefmark{2}Department of Artificial Intelligence and Machine Learning, 
University School of Automation and \\Robotics, Guru Gobind Singh Indraprastha University, Delhi, India}
\IEEEauthorblockA{\IEEEauthorrefmark{3}Department of Electronics and Communication Engineering, 
Indira Gandhi Delhi Technical \\University for Women, Delhi, India}
\IEEEauthorblockA{\IEEEauthorrefmark{4}Department of Gastroenterology and HNU, 
All India Institute of Medical Sciences Delhi, India}
}

\title{WCEBleedGen: A wireless capsule endoscopy dataset and its benchmarking for automatic bleeding classification, detection, and segmentation}

\begin{document}

\maketitle

\begin{abstract}
\textit{Objective:} Computer-based analysis of Wireless Capsule Endoscopy (WCE) is crucial. However, a medically annotated WCE dataset for training and evaluation of automatic classification, detection, and segmentation of bleeding and non-bleeding frames is currently lacking. \textit{Methods:} The present work focused on development of a medically annotated WCE dataset called WCEbleedGen for automatic classification, detection, and segmentation of bleeding and non-bleeding frames. It comprises 2,618 WCE bleeding and non-bleeding frames which were collected from various internet resources and existing WCE datasets. A comprehensive benchmarking and evaluation of the developed dataset was done using nine classification-based, three detection-based, and three segmentation-based deep learning models. \textit{Results:} The dataset is of high-quality, is class-balanced and contains single and multiple bleeding sites. Overall, our standard benchmark results show that Visual Geometric Group (VGG) 19, You Only Look Once version 8 nano (YOLOv8n), and Link network (Linknet) performed best in automatic  classification, detection, and segmentation-based evaluations, respectively. \textit{Conclusion and significance:} Automatic bleeding diagnosis is crucial for WCE video interpretations. This diverse dataset will aid in developing of real-time, multi-task learning-based innovative solutions for automatic bleeding diagnosis in WCE. The dataset and code are publicly available here \footnote{\href{https://zenodo.org/records/10156571}{https://zenodo.org/records/10156571}} \footnote{\href{https://github.com/misahub2023/Benchmarking-Codes-of-the-WCEBleedGen-dataset}{https://github.com/misahub2023/Benchmarking-Codes-of-the-WCEBleedGen-dataset}}.
\end{abstract}

\begin{IEEEkeywords}
Open-source datasets, Gastrointestinal bleeding, Wireless Capsule Endoscopy, Classification, Detection, Segmentation
\end{IEEEkeywords}

\section{Introduction}
\label{intro}

Gastrointestinal (GI) bleeding is the most common GI diagnosis in both inpatient and emergency settings \cite{wilcox2009mortality, b1}. It results in 400,000 hospital admissions and 300,000 deaths per year globally, with mortality rates between 5 and 10 percent \cite{gralnek2008management}. It has been identified as the most common GI cause of 30-day readmission in U.S. hospitals during 2014-15 \cite{peery2019burden}. Locating the source of GI bleeding is crucial for timely treatment initiation.

The most common sources of GI bleeding are the upper and lower GI tracts. However, in approximately between 5 and 10 percent of patients, the source of GI bleeding is the small bowel, which is difficult to examine using endoscopic methods due to its location, length, and excessive mobility \cite{gunjan2014small}. For GI bleeding within the small bowel, Wireless Capsule Endoscopy (WCE) is the state-of-the-art procedure with high sensitivity for detecting causes of bleeding \cite{nadler2014role}. WCE has also proven to be a useful tool for diagnosing bleeding, tumors, and inflammatory disease in other segments of the GI tract, such as the colon and stomach \cite{kitiyakara2005non}.

In WCE, a disposable capsule-shaped device travels inside the GI tract via peristalsis and comprises an optical dome, a battery, an illuminator, an imaging sensor, and a radio-frequency transmitter. During $8-12$ hours of WCE, a video of the GI tract transit is recorded on a device attached to the patient’s belt which produces about $57,000-100,000$ frames that are subsequently analyzed visually by subspecialized gastroenterologists \cite{thakur2024vce}. With two to three hours of reading time per WCE for a subspecialized gastroenterologist, this visual evaluation is not only time consuming but also suffers from substantial heterogeneity and sub-optimal inter- and intra-observer disagreement \cite{cortegoso2022inter}.

Due to an increasing lack of subspecialized gastroenterologists worldwide and a need for robust and repeatable WCE analysis, there is an unmet need for developing robust, interpretable and generalized Artificial Intelligence (AI) models that can decrease the workload of gastroenterologists by providing computer-aided classification, detection, and segmentation of bl\-eeding and non-bleeding frames in WCE.

Research on streamlining multi-task learning in automatic bleeding diagnosis in WCE i.e., the classification of bleeding and non-bleeding frames in WCE, and further detection, and segmentation of bleeding in WCE has been hindered by the absence of dedicated datasets. Table \ref{tab1} provides an overview of existing datasets in this domain. These datasets are characterized by imbalanced classes and do not adequately address the challenges of this AI problem such as the need for frames with multiple bleeding sites, and different types of bleeding shades,  mixed with intestinal fluids, chyme, etc. The frames within these existing datasets do not exhibit diverse dimensions and thus necessitate pre-processing prior to utilization of any machine learning or deep learning models. None of the existing datasets include a combination of class labels, bounding boxes, and medical markings which could promote multi-task learning-based research in this field.

\begin{table*}[htbp]
\caption{Existing annotated/partiality annotated  datasets of wireless capsule endoscopy. NR= not reported.}
\centering
\begin{adjustbox}{width=1.8\columnwidth}
\begin{tabular}{ccccccccc}
\hline
Dataset name & Year& Size & No. of bleeding& No. of non-& Dimension& Cross-site & Contains non-bleeding& Labels \\
 & &  & frames &bleeding frames& & variability & frames or not&  \\
\hline

KID (multiple anomalies) &  2014 & 335 MB & 5 &NR& $320\times320$ & No & No & Class labels, binary masks \\
\hline

Farah deeba (only bleeding) &  2016 & 1.04 MB & 50& 0& $568\times430$ & No & No & Binary masks\\
\hline

Set 1 (red lesions) and Set &  2017 & 506 MB & NR &NR & $320\times320$,  & Yes & Yes & Binary masks (red \\
2 (bleeding and non-bleeding)&  &  &&  & $512\times512$ &  &  & lesions only)  \\
MICCAI 2017&  &  &  &  &  &  &  \\
\hline

EndoSLAM dataset & 2020 & 23.20 GB & 0 & 0 &$256\times256$ & No & No & NR\\
\hline

Kvasir capsule (multiple anomalies)& 2021 & 89.00 GB & 458& NR & $336\times336$ & No & Yes &  Class labels, Bounding boxes\\
\hline

\textbf{WCEBleedGen dataset}  & \textbf{2023} & \textbf{168 MB} & \textbf{1309} & \textbf{1309} &\textbf{$224\times224$}& \textbf{Yes} & \textbf{Yes} & \textbf{Class labels, binary}  \\
\textbf{(bleeding and non-bleeding)} & &&&&&&& \textbf{masks, and bounding box}\\
\hline
\end{tabular}
\end{adjustbox}
\label{tab1}
\end{table*}

Due to the limited availability of dedicated WCE datasets that specifically classify bleeding as a label, studies have incorporated various lesions such as angioectasia, ulcers, and polyps to develop automated AI models. While these lesions can lead to bleeding, the bleeding itself may not be directly observable in the WCE frames \cite{b3, b2}. Consequently, AI models developed using these datasets, cannot be employed on real-world data. Thus, for a robust, interpretable and generalized AI development in this domain, a dedicated dataset which only contains bleeding and non-bleeding WCE frames is essential. 

The present work introduces WCEbleedGen, an open-source dataset comprising a total of $2,618$ WCE frames collected from various internet repositories and existing datasets. The dataset consists of high resolution frames of uniform size, equally divided into the two classes namely bleeding and non-bleeding. Additionally, the dataset has been meticulously validated by subspecialized gastroenterologists, ensuring high accuracy. It features comprehensive annotations, including class labels, manually generated binary masks, and precise bounding boxes, making it a robust resource for advanced analysis. These features facilitate the development of combined solutions for the classification, detection, and segmentation of bleeding regions in WCE frames.

Table \ref{tab2} presents an overview of the recent advancements in AI techniques developed for automating the analysis of bleeding regions in WCE frames. Although these research works have made valuable contributions, they are subject to certain constraints. The metrics obtained in these works pose challenges in terms of validation and comparison due to the scarcity of data and the lack of established benchmarks. In addition, none of the previous studies report results for classification, segmentation, and detection simultaneously. Addressing these gaps, the present work also encompasses comprehensive benchmarking and evaluation of the developed dataset using nine conventional classification models, three detection models, and three segmentation models employing standard performance metrics. This establishes a performance standard for researchers to use in comparing and assessing their own work.

\begin{table}[htbp]
\centering 
\caption{Recent AI works in separate detection, classification, and segmentation of bleeding regions in WCE (sorted by publication year).}
\label{tab2}
\begin{adjustbox}{width=\columnwidth}
\begin{tabular}{lllll}
\hline
Author and Ref. & Year & Accuracy & Dataset & Task Performed \\
\hline 
Cahrfi et al.~\cite{charfi2024abnormalities} & 2024 & 99.90\%   & MICCAI 2017 & Detection \\
Amiri et al.~\cite{amiri2024combining}   & 2024 & 97.82\%    & Kvasir, GIANA 2017, & Detection \\
  &&    &  Red-lesion endoscopy, & \\
 &&    &  KID & \\
Lafraxo et al.~\cite{lafraxo2024computer} & 2024 & 99.39\% & MICCAI 2017 & Detection \\
Kaur et al.~\cite{kaur2023wireless} & 2023 & 98.75\% & WCE Curated Colon & Classification \\
 &&    &  Disease Dataset & \\
Padmavathi et al.~\cite{padmavathi2023wireless} & 2023 & 99.60\% & Private Dataset & Detection \\
Sreejesh et al.~\cite{sreejesh2023bleeding} & 2023 & 95.75\% & Private Dataset & Classification \\
Padmavathi et al.~\cite{padmavathi2023effective} & 2023 & 99.12\% & Kvasir V2 & Segmentation \\
Lafraxo et al.~\cite{lafraxo2023semantic} & 2023 & 99.16\% & MICCAI 2017 & Segmentation \\
Bordbar et al.~\cite{bordbar2023wireless} & 2023 & 99.20\% & Private Dataset & Classification\\
Naz et al. ~\cite{naz2023comparative} & 2023 & 99.24\% & Kvasir & Classification\\
Singh et al. ~\cite{singh2022explainable} & 2022 & 91.92\% & Private Dataset & Classification\\
Mohankumar et al. ~\cite{mohankumar2022rnn} & 2022 & 93.27\% & Private Dataset & Classification\\

\hline
\end{tabular}
\end{adjustbox}
\end{table}

\section{Methodology}

The objective of this work was to develop and benchmark a WCE dataset consisting of bleeding and non-bleeding frames. The dataset development process is detailed in sub-section \ref{ww}. The information related to pre-existing datasets is detailed in sub-section \ref{aa}. To  benchmark and validate the effectiveness of the developed dataset, nine classification-based deep learning models, three detection-based deep learning models, and three segmentation-based deep learning models were evaluated for automatic classification, detection, and segmentation of bleeding and non-bleeding frames separately. This evaluation is described in sub-section \ref{problem definition}. The implementation settings are mentioned in sub-section \ref{exp}.

\subsection{Data Collection, Processing and its Annotation} \label{ww}

\begin{figure}[htbp]
\centering\includegraphics[width=0.8\columnwidth]{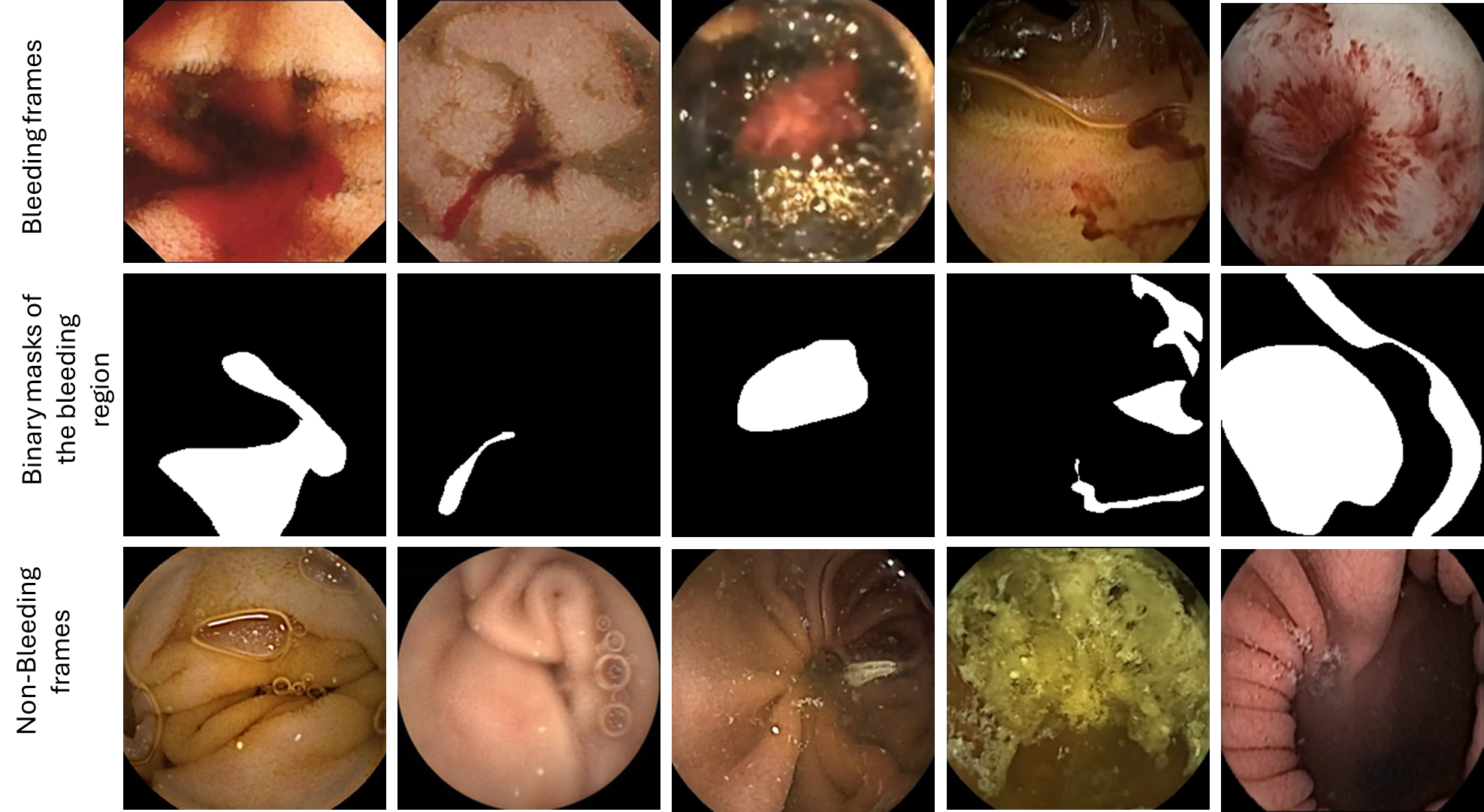}
\caption{Sample bleeding and non-bleeding frames and their binary masks present in the WCEBleedGen dataset. 
} \label{fig1}
\end{figure}

The first phase of the dataset development involved the compilation and extraction of frames depicting both bleeding and non-bleeding instances from multiple sourc\-es including various internet repositories and pre-existi\-ng  datasets. Table \ref{tab3} presents the sources for each frame in the dataset, to provide transparency regarding the origin of the data. A total of $2,618$ frames were collected, with the bleeding and non-bleeding classes consisting of $1,309$ frames each to ensure class balance. Sample frames of the dataset are shown in Figure~\ref{fig1}. 

After data collection, the frames were resized to a consistent size of $224\times224$ pixels, which corresponds to the default input size for common transfer learning techniques \cite{deng2009imagenet}. Then the binary masks were manually generated for all the frames followed by development of bounding boxes through an automated python script to ensure accurate detection and segmentation of bleeding regions and to obtain a comprehensive understanding of medical features. Additionally, the dataset was manually validated by subspecialized gastroenterologists to ensure correctness and reliability in clinical interpretation.

The data was then organized into a folder structure, with separate folders for bleeding and non-bleeding classes, as well as distinct folders for binary masks, images, and bounding box markings within each class folder. Three formats of bounding box markings were included inside the bounding box folder namely the text, extensible markup language (XML), and text markings compatible with You Only Look Once (YOLO) models.

\begin{table}[htbp]
    \centering
     \caption{WCEBleedGen dataset details.}
        \begin{adjustbox}{width=\columnwidth}
    \begin{tabular}{lllll}
    \hline
      Type of data &  Original source  &	Image S.No. &	Total No. &	Binary mask \\
      &&&&availability (previously)\\
\hline
& 1. Set-2 data~\cite{redlesion} &	1-152 &	152	& yes \\	
 & 2. farah deeba data~\cite{deeba2016automated} &	153-202	 & 50 & 	yes \\
& 3. KID~\cite{koulaouzidis2017kid} &	203-207	 & 5 & 	yes \\
Bleeding & 4. kvasir capsule (fresh blood)~\cite{smedsrud2021kvasir}&	208-650	& 443 &	no, matrix given \\
&  5. kvasir capsule (hematin blood)~\cite{smedsrud2021kvasir} &	651-660	&10	&no, matrix given \\
& 6. Self collected (YouTube video 1 and 2) &	661-952 &	292 &	no \\
& 7. Gastrolab &	953-1309 &	357 &	no
\\
\hline
& Total bleeding images and annotations =& & 1309 & \\
         \hline 
      &  1. Set-1 data~\cite{redlesion}&	1-500 &	500	 & no \\
 Non-bleeding   & 2. Set-2 data~\cite{redlesion}	& 501-580 & 	80	& no \\
 &3. Self collected (YouTube video)&	581-761	 & 181 &	no \\
 &4. Gastrolab&	762-1309	 & 548 &	no \\

\hline
& Total non-bleeding images and annotations =& & 1309 & \\
         \hline 

    \end{tabular}
      \end{adjustbox}
\label{tab3}
\end{table}

\subsection{Pre-existing Datasets}\label{aa}

\subsubsection{\texorpdfstring{KID~\cite{koulaouzidis2017kid}}{KID}}
KID is the first dataset used in WCE-AI research, featuring over $2,500$ WCE frames with a size of $320\times320$ pixels. It includes class labels and binary masks, enabling both classification and segmentation tasks. The frames are divided into $5$ categories: normal, vascular lesions with angioectasias (including bleeding), inflammatory lesions (such as mucosal aphthae and ulcers, erythema, cobblestoning, and luminal stenosis), lymphangiectasias, and polypoid lesions. The data was collected from $6$ different centres and the frames were acquired using a MiroCam (IntroMedic Co, Seoul, Korea) capsule. The KID dataset has the limitation of unbalanced data, with certain classes — particularly bleeding — having significantly more frames compared to the normal class. This imbalance poses a challenge for classification tasks. The dataset was previously available here\footnote{\href{https://mdss.uth.gr/datasets/endoscopy/kid/}{https://mdss.uth.gr/datasets/endoscopy/kid/}}.

\subsubsection{\texorpdfstring{Farah Deeba Data~\cite{deeba2016automated}}{Farah Deeba Data}}
This dataset was developed in $2016$, as a part of a research work in ~\cite{deeba2016automated}. It contains 50 frames with a size of $568\times430$ pixels along with their ground truth in the form of binary masks. The dataset was previously available here\footnote{\href{https://sites.google.com/site/farahdeeba073/Research/resources}{https://sites.google.com/site/farahdeeba073/Research/resources}}.
\subsubsection{\texorpdfstring{Red Lesion Endoscopy Dataset~\cite{redlesion, coelho2018deep}}{Red Lesion Endoscopy Dataset}}
This dataset was released in $2017$ and contains a total of $3,895$ WCE frames divided into $2$ sets. Set $1$ contains $3,295$ non-sequential frames with a size of $512\times512$ pixels containing red lesions, each accompanied by its binary mask. Whereas, set $2$ contains $600$ sequential frames with a size of $320\times320$ pixels including both bleeding and non-bleeding classes. The dataset is freely available here\footnote{\href{https://rdm.inesctec.pt/dataset/nis-2018-003}{https://rdm.inesctec.pt/dataset/nis-2018-003}}.
\subsubsection{\texorpdfstring{Kvasir-Capsule~\cite{smedsrud2021kvasir}}{Kvasir-Capsule}}
Kvasir-Capsule database is the largest dataset available for WCE-AI research since $2020$. It was collected at the Norwegian Hospital and consists of $4,741,504$ frames from about $117$ videos, including both normal and abnormal classes. The Olympus ECS10 endocapsule was utilized for video frame extraction with a size of $336\times336$ pixels. Among the $117$ videos, $74$ videos are unlabelled which comprise of $4,741,504$ frames, and may be utilized for unsupervised learning-based research. There are $47,238$ labeled frames with medically verified bounding boxes and $14$ class labels that may be utilized for development of segmentation and detection tasks. The data may be downloaded from here\footnote{\href{https://osf.io/dv2ag/}{https://osf.io/dv2ag/}}.

\subsection{Bio-mathematical problem definition and evaluation metrics}\label{problem definition}
Let \begin{math}
\mathnormal{D} = \{(X_i, y_i) |       i = 1, ... N\}
\end{math} represent the developed dataset, where $N$ = $2618$ is the total number of frames in the dataset. \begin{math} X_i \in \mathnormal{R} ^{224 \times 224 \times 3} \end{math} is the $i_{th}$ frame and \begin{math}  y_i \in \{0,1\} \end{math} is the binary label indicating non-bleeding (0) or bleeding (1). The research problem is defined for three tasks:

First, the classification task where a binary classifier $f$, produces an output probability score $p_i$ such that: 

\begin{math}
\hat{y_i} = 
\biggl\{ 
    \begin{aligned}
     & 1 \quad & if \quad p_i \geq 0.5 \\
     &  0 \quad & otherwise
\end{aligned}
\end{math}

where \begin{math} p_i = f (X_i).
 \end{math}

The classifiers included Residual Network 101 Version 2 (ResNet101V2) \cite{he2016identity}, Inception Version 3 (InceptionV3) \cite{nolan2010inception}, Densely Connected Convolutional Network 169 (DenseNet169) \cite{huang2017densely}, Convolutional Neural Network with NeXt Steps (ConvNeXT) \cite{liu2022convnet}, Extreme Inception (Xception) \cite{chollet2017xception}, Visual Geometry Group 19 (VGG19) \cite{simonyan2013deep}, Neural Architecture Search Network Mobile (NasNetMobile) \cite{zoph2018learning}, Mobile Network (MobileNet) \cite{howard2017mobilenets}, and Inception Residual Network Version 2 (Inception ResNetV2) \cite{simonyan2014very}.

Second, the detection task where a detection model $g$ predicts a set of bounding boxes $\{(x_k, y_k, w_k, h_k)\}$ where $k$ indexes the detected bleeding regions. Third, the segmentation task where a segmentation model $h$ predicts a binary mask \begin{math} M_i \in \{0,1\}^{224\times224} \end{math} where each pixel value indicates the presence (1) or absence (0) of bleeding. For detection tasks, we utilized three versions of YOLO models \cite{redmon2016you} namely YOLO Version 5 (YOLOv5), YOLO Version 8 nano (YOLOv8n), and YOLO Version 8 extra large (YOLOv8x). In the segmentation tasks, we employed the Unet \cite{ronneberger2015u}, segmentation network (Segnet) \cite{badrinarayanan2015segnet}, and Link network (Linknet)  \cite{chaurasia2017linknet} architectures. The utilized detection and segmentation networks have shown excellent performance in previous works \cite{kora2022transfer, alzubaidi2021novel, qureshi2023comprehensive, yin2022u}, and hence we selected them as standard models for benchmarking.

\subsection{Implementation settings}\label{exp}

The models were implemented using Python scripts with TensorFlow as the backend on a high-performance super-computer consisting of four port 40 GB DGX A100 NVIDIA workstations, running on a pre-installed Ubuntu Linux-based operating system. Additionally, the image labeler MATLAB toolbox was utilized for generating precise binary mask annotations for all the $1,309$ bleeding frames. For some of the frames available in the Red Lesion Endoscopy Dataset (Set 2), direct rectangle boxes were drawn as binary mask annotations. The bounding boxes were generated using the binary masks with the help of a python script using libraries like pandas, and OpenCV. The script has been provided on our github.

For classification tasks, all the models were compiled with Adam optimizer and categorical cross-entropy loss with the default learning rate as $0.001$. The batch size for training was kept as $32$. On-the-fly data augmentations like rotation (up to $\pm10$ degrees), width and height shifting (up to $\pm0.1$), zoom (up to $\pm0.1$), horizontal flipping, and normalization to $[0,1]$ were done only to increase the quantum of the data for classification tasks. All classification models were utilized with pre-trained weights from imagenet dataset and detection models were loaded with weights from the Common Objects in Context (COCO) dataset~\cite{deng2009imagenet}. For detection, the Adam optimizer was utilized with learning rate and momentum set to default which was automatically determined as $0.002$ and $0.9$ respectively. The batch size for training was kept as 16. All detection models were setup using the in-built functions available in the Ultralytics library.  For segmentation, Adam optimizer was used with $0.0001$ learning rate, binary cross-entropy loss and a batch size of $32$. The models were built using keras library. No separate on-the-fly augmentations were done for the detection and segmentation tasks.

For classification tasks, averaged accuracy, accuracy on the last epoch, averaged loss, and loss on the last epoch were considered for the training and validation data, while precision, recall, and F1 score were considered for the test set evaluation. For detection tasks, Box Loss (BL), Classification Loss (CLS), and Distribution Focal Loss (DFL) were used for training data, while precision, recall, and mean Average Precision (mAP) were considered for test dataset evaluation. For the validation data, BL, CLS,  mAP, recall, and precision were reported. The evaluation of the segmentation tasks for the training, validation, and test dataset were based on accuracy, precision, recall, dice coefficient, loss, accuracy, and Intersection over Union (IoU).

For evaluation of the dataset for classification, segmentation and detection tasks, the dataset was randomly split into $70-20-10$ ratio, where $70\%$ was used for training, $20\%$ was used for validation and the $10\%$ subset was used for testing purposes. A fixed random seed was initialized in all the scripts to maintain reproducibility of the scripts and to ensure consistency. All models for classification, detection, and segmentation tasks were trained for $250$ epochs without any modification, parameter sharing or hyper parameter tuning. Table \ref{tab4} depicts the total parameters including both trainable and non-trainable along with their respective sizes for all of the AI models utilized in this work.

\begin{table}[htbp]
\centering
\caption{Details of the utilized artificial intelligence models.}
\begin{adjustbox}{width=\columnwidth}
\begin{tabular}{cccccc}
\hline
S. No. & Transfer learning algorithm & Trainable parameters & Non-trainable parameters & Total parameters & Size \\
\hline

\multicolumn{6}{c}{(a) Classification}  \\ \hline
1. & ConvNeXt & 0.10 M & 87.60 M & 87.70 M & 335.10 MB \\
2. & VGG19 & 0.05 M & 20.00 M & 20.10 M & 76.60 MB \\
3. & MobileNetV2 & 0.13 M & 2.30 M & 2.40 M & 9.50 MB \\
4. & DenseNet169 & 0.16 M & 12.60 M & 12.80 M & 50.10 MB \\
5. & Inception ResNetV2 & 0.08 M & 54.30 M & 54.40 M & 209.2 MB \\
6. & InceptionV3 & 0.10 M & 21.80 M & 21.90 M & 84.30 MB \\
7. & NasNetMobile & 0.10 M & 4.30 M & 4.40 M & 18.30 MB \\
8. & Xception & 0.20 M & 20.90 M & 21.10 M & 80.70 MB \\
9. & ResNet101V2 & 0.20 M & 42.60 M & 42.80 M & 164.20 MB \\
\hline
\multicolumn{6}{c}{(b) Detection}  \\ \hline
1. & YOLOv5nu & - & - & 2.50 M & 5.04 MB  \\
2. & YOLOv8n & - & - & 3.00 M & 5.97 MB \\
3. & YOLOv8x & - & - & 68.20 M & 130 MB \\
\hline
\multicolumn{6}{c}{(c) Segmentation}  \\ \hline
1. & Linknet & 7.40 M & 0.009 M & 7.40 M & 28.60 MB \\
2. & Segnet & 33.40 M & 0.02 M & 33.40 M & 127 MB \\
3. & Unet & 0.41 M & 0.002 M & 0.41 M & 1.73 MB \\
\hline

\end{tabular}
\label{tab4}
\end{adjustbox}
\end{table}

\section{Results}
The results have been discussed as per the three tasks namely the classification, detection, and segmentation performed on the proposed datasets. Table \ref{tab5} and \ref{tab6} detail the achieved performance metrics for classification task on training, validation and test set respectively. The training and validation performance metrics for detection tasks have been detailed in Table \ref{tab7}. Table \ref{tab8} detail the results achieved upon testing the detection-based AI models. Table \ref{tab9} and \ref{tab10} summarize the achieved evaluation metrics on training, validation and test set for LinkNet, SegNet and UNet segmentation models.

\subsection{Classification task}

Over the average of $250$ epochs, a highest accuracy and lowest loss of up to $0.99$ and $0.004$ respectively were observed for the training dataset, whereas the highest and the lowest values for accuracy and loss on last epoch were found to be $1$ and $3.25\times 10^{-10}$ respectively. The highest accuracy of $0.78$ was recorded for the validation data, both as an average over $250$ epochs and in the last epoch. This was accompanied by the lowest average loss of $6.46$ and the lowest last epoch loss of $6.93$ for the validation data. For the test set, the highest values achieved for the macro averages of precision, recall, and F1-score were $0.63$, $0.63$, and $0.63$ respectively. Likewise, the highest value for weighted average of precision, recall and F1-score was also $0.63$. For the classification task nearly all AI models achieved notably high accuracy ranging between $0.96-1$, with minimal loss values on the training data. 

\begin{table*}[htbp]
\centering
\caption{Performance metrics for the classification of bleeding and non-bleeding frames in the training and validation data of the WCEBleedGen data.}
\begin{adjustbox}{width=2\columnwidth} \label{tab5}
\begin{tabular}{lllllllllll}
\hline
\multicolumn{1}{l}{Type of data}     & Evaluation Metrics               & ResNet101V2 & InceptionV3 & DenseNet169 & ConvNeXT & Xception & VGG19    & NasNetMobile & MobileNet & Inception ResNetV2    \\

\hline

\multirow{4}{*}{For training data}   & Accuracy (average of 250 epochs) & 0.99        & 0.97        & 0.99        & 0.95     & 0.99     & 0.99     & 0.99         & 0.99      & 0.99                      \\
                                     & Accuracy (last epoch)            & 1           & 0.99        & 1           & 0.96     & 1        & 1        & 1            & 1         & 1                            \\
                                     & Loss (average of 250 epochs)     & 0.014       & 0.089       & 0.009       & 0.18     & 0.004    & 0.008    & 0.006        & 0.004     & 0.005                    \\
                                     & Loss (last epoch)                & 3.25e-10    & 0.009       & 4.85e-08    & 0.15     & 3.16e-08 & 2.70e-06 & 1.45e-07     & 6.77e-09  & 8.73e-08                 \\ \hline
\multirow{4}{*}{For validation data} & Accuracy (average of 250 epochs) & 0.47        & 0.50        & 0.50        & 0.58     & 0.55     & 0.78     & 0.48         & 0.52      & 0.52                      \\  
                                     & Accuracy (last epoch)            & 0.47        & 0.50        & 0.50        & 0.57     & 0.56     & 0.78     & 0.48         & 0.52      & 0.52                \\
                                     & Loss (average of 250 epochs)     & 636.05      & 84.39       & 44.81       & 10.37    & 112.91   & 30.07    & 36.50        & 6.46      & 223.93                   \\
                                     & Loss (last epoch)                & 637.06      & 130.15      & 46          & 16.71    & 107.6    & 43.98    & 46.72        & 6.93      & 209.77                  \\ \hline
\end{tabular}
\end{adjustbox}
\end{table*}

\begin{table*}[htbp]
\centering 
\caption{Performance metrics for the classification of bleeding and non-bleeding frames in the test set of the WCEBleedGen dataset.}
\begin{adjustbox}{width=2\columnwidth}\label{tab6}
\begin{tabular}{llllllllll}
\hline
Evaluation Metrics           & ResNet101V2 & InceptionV3 & DenseNet169 & ConvNeXt & Xception & VGG19 & NasNetMobile & MobileNetV2 & Inception ResNetV2  \\ \hline
Precision (macro average)    & 0.52        & 0.63        & 0.25        & 0.37    & 0.57     & 0.63  & 0.28         & 0.25        & 0.37                      \\
Precision (weighted average) & 0.52        & 0.63        & 0.25        & 0.37    & 0.57     & 0.63  & 0.28         & 0.25        & 0.37                   \\
Recall (macro average)       & 0.52        & 0.52        & 0.50        & 0.38    & 0.53     & 0.60  & 0.40         & 0.50        & 0.39                    \\
Recall (weighted average)    & 0.52        & 0.52        & 0.50        & 0.38    & 0.51     & 0.60  & 0.40         & 0.50        & 0.39                    \\
F1-score (macro average)     & 0.51        & 0.38        & 0.33        & 0.36    & 0.43     & 0.58  & 0.31         & 0.33        & 0.36                     \\
F1-score (weighted average)  & 0.51        & 0.38        & 0.33        & 0.36    & 0.42     & 0.58  & 0.31         & 0.33        & 0.36                       \\ \hline
\end{tabular}
\end{adjustbox}
\end{table*}

\subsection{Detection task}
On the training data, the lowest values for BL, CLS, and DFL averaged over $250$ epochs were found to be up to $1.15$, $1.22$, and $1.44$, respectively. Similarly, $0.80$, $0.66$, and $1.19$ were obtained as the lowest BL, CLS and DFL values respectively, for the last epoch. The highest precision, recall, mAP@50, and mAP@95 on the validation set, averaged over $250$ epochs, were found to be up to $0.62$, $0.55$, $0.57$, and $0.28$, respectively. These were accompanied by lowest average BL, CLS and DFL values of up to $1.73$, $1.68$, and $1.99$ respectively. Similarly for the last epoch, $0.69$, $0.62$, $0.65$, and $0.33$ were obtained as the highest values for precision, recall, mAP@50, and mAP@95 respectively. The lowest values for BL, CLS, and DFL were observed at the last epoch, reaching $1.67$, $1.40$, and $2.04$, respectively. For the $10\%$ test set, the highest values for precision, recall, mAP@50, and mAP@95 were $0.69$, $0.64$, $0.63$, and $0.36$, respectively.

\begin{table}[htbp]
\centering
\caption{Performance metrics for the detection of bleeding region in the training and validation data of the WCEBleedGen dataset.}
\begin{adjustbox}{width=\columnwidth} 
\label{tab7}
\begin{tabular}{llllllllllll}
\hline
Type of Data & Evaluation Metrics & YOLOv5nu & YOLOv8n & YOLOv8x & \\ \hline
\multirow{6}{*}{For Training Data} & Box Loss (Avg of 250 epochs) &1.19 &1.15 &1.26 & \\
& Classification Loss (Avg of 250 epochs) &1.27 &1.22 &1.45 & \\
& Distribution Focal Loss (Avg of 250 epochs) &1.46 &1.44 &1.58 & \\
& Box Loss (Last epoch) &0.84 &0.80 &0.81 & \\
& Classification Loss (Last epoch) &0.67 &0.66 &0.72 & \\
& Distribution Focal Loss (Last epoch) &1.22 &1.19 &1.26 & \\ \hline
\multirow{12}{*}{For Validation Data} & Box Loss (Avg of 250 epochs) &1.77 &1.79 &1.73 & \\
& Classification Loss (Avg of 250 epochs) &1.70 &1.68 &1.78 & \\
& Distribution Focal Loss (Avg of 250 epochs) &1.99 &2.03 &3.73 & \\
& Box Loss (Last epoch) &1.80 &1.84 &1.67 & \\
& Classification Loss (Last epoch) &1.40 &1.44 &1.48 & \\
& Distribution Focal Loss (Last epoch) &2.05 &2.18 &2.04 & \\
& Precision (Avg of 250 epochs) &0.61 &0.62 &0.54 & \\
& Recall (Avg of 250 epochs) &0.54 &0.55 &0.49 & \\
& Mean Average Precision@50 (Avg of 250 epochs) &0.56 &0.57 &0.49 & \\
& Mean Average Precision@95 (Avg of 250 epochs) &0.28 &0.28 &0.25 & \\
& Precision (Last epoch) &0.69 &0.67 &0.65 & \\
& Recall (Last epoch) &0.62 &0.62 &0.59 & \\
& Mean Average Precision@50 (Last epoch) &0.64 &0.65 &0.61 & \\
& Mean Average Precision@95 (Last epoch) &0.33 &0.33 &0.33 & \\ \hline
\end{tabular}
\end{adjustbox}
\end{table}

\begin{table}[htbp]
\centering 
\caption{Performance metrics for the detection of bleeding region in the test set of the WCEBleedGen dataset.}
\label{tab8}
\begin{adjustbox}{width=1\columnwidth}
\begin{tabular}{llll}
\hline
Evaluation Metrics        & YOLOv5nu & YOLOv8n & YOLOv8x \\ \hline
Precision                 & 0.68    & 0.69   & 0.65   \\
Recall                    & 0.64    & 0.56   & 0.59   \\
Mean Average Precision@50 & 0.63    & 0.63   & 0.59   \\
Mean Average Precision@95 & 0.35     & 0.36    & 0.30     \\ \hline
\end{tabular}
\end{adjustbox}
\end{table}
\subsection{Segmentation task}
The highest values for accuracy, dice coefficient, IoU, precision and recall averaged over 250 epochs were obtained as $0.99$, $0.93$, $0.89$, $0.96$, and $0.99$ respectively, accompanied by lowest average loss value of $0.01$ on the training data. Similarly $0.99$, $0.98$, $0.97$, $0.99$, and $0.99$ were obtained as the highest accuracy, dice coefficient, IoU, precision and recall respectively, for the last epoch with the lowest loss of $0.004$. For the validation set, $0.97$, $0.87$, $0.80$, $0.89$, and $0.90$ were achieved as the highest accuracy, dice coefficient, IoU, precision and recall respectively, with the lowest loss value of up to $0.06$, averaged over $250$ epochs. Likewise for the last epoch, the highest values for accuracy, dice coefficient, IoU, precision and recall were achieved up to $0.98$, $0.95$, $0.90$, $0.93$, and $0.97$ respectively, along with lowest loss of $0.04$. Whereas for the test set, $0.99$, $0.98$, $0.95$, $0.90$, and $0.94$ were obtained as the highest accuracy, precision, recall, IoU and dice coefficient respectively.

\begin{table}[htbp]
\centering
\caption{Performance metrics for the segmentation of bleeding region in the training and validation data of the WCEBleedGen dataset.} \label{tab9}
\begin{adjustbox}{width=\columnwidth}
\begin{tabular}{llllllllllll}
\hline
Type of Data & Evaluation Metrics & Linknet & Segnet & Unet & \\ \hline
\multirow{12}{*}{For Training Data} 
& Accuracy (Avg of 250 epochs) &0.99 &0.98 &0.98 & \\
& Dice Coefficient (Avg of 250 epochs) &0.93 &0.46 &0.83 & \\
& Intersection over Union (Avg of 250 epochs) &0.89 &0.31 &0.73 & \\
& Loss (Avg of 250 epochs) &0.01 &0.27 &0.05 & \\
& Precision (Avg of 250 epochs) &0.96 &0.91 &0.92 & \\
& Recall (Avg of 250 epochs) &0.99 &0.97 &0.92 & \\
& Accuracy (Last epoch) &0.99 &0.99 &0.99 & \\
& Dice Coefficient (Last epoch) &0.98 &0.64 &0.95 & \\
& Intersection over Union (Last epoch) &0.97 &0.47 &0.90 & \\
& Loss (Last epoch) &0.004 &0.11 &0.01 & \\
& Precision (Last epoch) &0.99 &0.98 &0.96 & \\
& Recall (Last epoch) &0.99 &0.98 &0.96 & \\ \hline
\multirow{12}{*}{For Validation Data} 
& Accuracy (Avg of 250 epochs) &0.97 &0.97 &0.96 & \\
& Dice Coefficient (Avg of 250 epochs) &0.87 &0.46 &0.77 & \\
& Intersection over Union (Avg of 250 epochs) &0.80 &0.31 &0.65 & \\
& Loss (Avg of 250 epochs) &0.06 &0.28 &0.10 & \\
& Precision (Avg of 250 epochs) &0.89 &0.88 &0.87 & \\
& Recall (Avg of 250 epochs) &0.90 &0.91 &0.82 & \\
& Accuracy (Last epoch) &0.98 &0.99 &0.97 & \\
& Dice Coefficient (Last epoch) &0.95 &0.65 &0.88 & \\
& Intersection over Union (Last epoch) &0.90 &0.48 &0.80 & \\
& Loss (Last epoch) &0.04 &0.11 &0.07 & \\
& Precision (Last epoch) &0.93 &0.95 &0.93 & \\
& Recall (Last epoch) &0.97 &0.98 &0.86 & \\ \hline
\end{tabular}
\end{adjustbox}
\end{table}

\begin{table}[htbp]
\centering 
\caption{Performance metrics for the segmentation of bleeding region in the test set of the WCEBleedGen dataset.}
\label{tab10}

\begin{tabular}{llll}
\hline
Evaluation Metrics        & Linknet & SegNet & Unet \\ \hline
Accuracy                  &0.98 &0.99 &0.97 \\
Recall                    &0.97 &0.98 &0.87 \\
Precision                 &0.92 &0.95 &0.93 \\
Intersection over Union   &0.90 &0.50 &0.79 \\
Dice Coefficient          &0.94 &0.65 &0.88 \\ \hline
\end{tabular}

\end{table}

\section{Discussion}

Manual interpretation of WCE frames with bleeding abnormality is a laborious, cost ineffective and error prone process due to the large number of frames captured during the capsule travelling throughout the GI tract. A subspecialized gastroenterologist spends at least two to three hours to evaluate a WCE video of a single patient~\cite{goel2022dilated, liu2009obscure, o2023capsule}. Thus to overcome this challenge, various AI models have been developed in the past to automate the process of bleeding analysis in WCE~\cite{haslach2023reading}. However, the development of integrated systems for bleeding analysis has been hindered by the lack of dedicated datasets. 

Our proposed dataset is the first of its kind which facilitates the development of a combined solution for performing simultaneous classification, detection and segmentation of bleeding regions in WCE frames. The proposed dataset consists of a total of $2,618$ frames with balanced classes of bleeding and non-bleeding and contains multiple types of abnormalities and lesions which contribute to bleeding. Furthermore, the dataset consists of high quality frames standardized to a uniform dimension of $224\times224$ pixels along with medical annotations in the form of binary masks and bounding boxes. It is freely available and can be used to build integrated solutions to perform automatic classification, detection, and segmentation of bleeding region in WCE frames. 

The proposed dataset was evaluated for classification, detection, and segmentation tasks using vanilla AI models to show the feasibility of AI in this field. The AI models used in this work were selected based on their promising performance in prior research works, particularly in classification, segmentation, and detection tasks on different biomedical datasets~\cite{kora2022transfer, alzubaidi2021novel, qureshi2023comprehensive, yin2022u}. All AI models were trained for $250$ epochs on a super-computer without any modifications. Common evaluation metrics were reported for each of the AI model.

VGG19, obtained an accuracy of up to $0.78$ on the validation data, performing better than the other models. This showed that VGG19 was successful in capturing the underlying patterns in the data, but not to an exceptionally high degree. Other models with very high training accuracy and significantly lower validation accuracy showed that the models were over-fit on some features present in the training data, leading to poor generalizability performance on validation data. All the models achieved low values for weighted average and macro average of precision, recall and F1-score on the test set. This showed that the models had poor generalization capability and could not be used in practical applications for unseen data without any modification. The overall poor performance of all models for the classification of bleeding frames showed the need of refinement of the AI models, hyper-parameter tuning or using ensemble learning along with other AI models.

There was very little variation in the performance of the three detection AI models with minimal difference in the achieved evaluation metrics. The models showed uniform performance across both the validation and test sets, indicating the absence of overfitting of the model on training or validation data. Yet, the attained metrics were not up to the mark, highlighting the model's inability to capture the complex patterns present within the dataset. Based on the metrics obtained in the testing phase, YOLOv8n architecture performed the best followed by YOLOv5nu and YOLOv8x. However, the overall low performance indicated the need of fine-tuning of the AI model to improve their performance for practical applications in clinical settings.

All three models in the segmentation task namely, the UNet, SegNet, and LinkNet, exhibited high accuracies accompanied with low loss values with LinkNet having the least loss on both training and validation data followed by UNet and SegNet. This indicated their ability to accurately classify pixels. SegNet achieved relatively lower values for IoU and dice coefficient on both training and validation data which shows that there was a lower amount of overlap between the predicted and the true annotations. It was inferred from the high values obtained for accuracy, precision and recall for SegNet, that the model was able to correctly classify the pixels, but was unable to define the boundary correctly leading to lower values for IoU and dice coefficient. A similar trend was observed in the IoU and dice coefficient obtained on the test set for SegNet. LinkNet and UNet, on the other hand, exhibit a favorable balance between the metrics observed for both the training and validation sets, along with a similar performance on the test set with LinkNet performing slightly better than UNet. This implied that the architectures were able to effectively learn the features of the dataset during the training process and have good generalization capability on unseen data as well. Further these AI models may be tweaked and fine tuned to achieve enhanced performance for real world applications.

Overall, the metrics obtained for classification and detection tasks were not up to the mark but demonstrated the potential of AI in this field with improvement and modification in the AI models. For the segmentation task, high metrics were obtained for the LinkNet model, thus further testing and tuning may be done for utilization in real world clinical setting. Furthermore, there is potential to enhance the evaluation metrics by incorporating image augmentation techniques and optimizing the model's hyper-parameters. 

The evaluation metrics presented in this work serve as valuable benchmarks for future researchers, enabling them to compare their findings, refine AI models, and address emerging challenges in the field. Baseline models help in establishing the foundation of the research problem by providing necessary context which in turn guides decision making process of researchers. They further expedite the development of highly efficient and accurate real-world systems, building upon the established foundation.

The present work is subject to certain limitations, which present opportunities for future research to address and overcome. The WCEBleedGen dataset consists of frames collected from diverse sources ranging from internet repositories to pre-existing datasets, thus it does not support per patient medical analysis as the actual source of the images is not known. Furthermore, the dataset does not enable longitudinal analysis of bleeding region. The variation of the bleeding region with time cannot be captured using the proposed dataset, due to different non-sequential image frames. The absence of sequential frames also hinders the recognition of the source of the blood flow inside the human body.

\section{Conclusion}

Automatic bleeding analysis in WCE interpretation enhances diagnostic accuracy, reduces clinician workload, and ensures timely detection of GI bleeding, improving patient outcomes. The present work demonstrated the potential of AI for automatic classification, detection, and segmentation of bleeding and non-bleeding frames in WCE. The proposed dataset, WCEBleedGen, validated by proficient gastroenterologists, enables the development of robust integrated solutions for multi-task learning in this field. The obtained evaluation metrics set a benchmark for future researchers for further improvements. The future directions of this research are geared towards the development of a larger dataset that encompasses multiple classes for more specific analysis of GI diseases.

\section*{Acknowledgment and Declarations}

P. Handa conceptualized the research idea, performed the data collection, mask analysis, literature review, and manuscript writing. M. Dhir performed the benchmarking and contributed in writing the initial draft of the manuscript. A. Mahbod, F. Schwarzhans, and R. Woitek were involved in suggestions and manuscript reviewing. N. Goel was involved in  project administration. D. Gunjan was involved in suggestions and performed the medical annotations.

The authors are thankful to Jyoti Dhatarwal for initial data collection, Harshita Mangotra for helping in development of the bounding boxes, Divyansh Nautiyal for correcting the multiple bleeding regions and re-making the bounding boxes, and Sanya, Shriya, and Sneha Singh for result replications on the GPU workstation and table entries. This dataset has been actively downloaded more than 1000 times and was utilized in Auto-WCEBleedGen Version 1 and 2 challenge as training dataset. 

The authors declare no personal, academic, or financial conflicts of interest associated with this work. No funding was received for this work.

\bibliographystyle{IEEEtran}
\bibliography{template/sample} %

\begin{thebibliography}{10}
\providecommand{\url}[1]{#1}
\csname url@samestyle\endcsname
\providecommand{\newblock}{\relax}
\providecommand{\bibinfo}[2]{#2}
\providecommand{\BIBentrySTDinterwordspacing}{\spaceskip=0pt\relax}
\providecommand{\BIBentryALTinterwordstretchfactor}{4}
\providecommand{\BIBentryALTinterwordspacing}{\spaceskip=\fontdimen2\font plus
\BIBentryALTinterwordstretchfactor\fontdimen3\font minus \fontdimen4\font\relax}
\providecommand{\BIBforeignlanguage}[2]{{%
\expandafter\ifx\csname l@#1\endcsname\relax
\typeout{** WARNING: IEEEtran.bst: No hyphenation pattern has been}%
\typeout{** loaded for the language `#1'. Using the pattern for}%
\typeout{** the default language instead.}%
\else
\language=\csname l@#1\endcsname
\fi
#2}}
\providecommand{\BIBdecl}{\relax}
\BIBdecl

\bibitem{wilcox2009mortality}
C.~M. Wilcox, B.~L. Cryer, H.~J. Henk, V.~Zarotsky, and G.~Zlateva, ``Mortality associated with gastrointestinal bleeding events: Comparing short-term clinical outcomes of patients hospitalized for upper gi bleeding and acute myocardial infarction in a us managed care setting,'' \emph{Clinical and Experimental Gastroenterology}, pp. 21--30, 2009.

\bibitem{b1}
D.~Lieberman, ``Gastrointestinal bleeding: initial management.'' \emph{Gastroenterology Clinics of North America}, vol.~22, no.~4, pp. 723--736, 1993.

\bibitem{gralnek2008management}
I.~M. Gralnek, A.~N. Barkun, and M.~Bardou, ``Management of acute bleeding from a peptic ulcer,'' \emph{New England Journal of Medicine}, vol. 359, no.~9, pp. 928--937, 2008.

\bibitem{peery2019burden}
A.~F. Peery, S.~D. Crockett, C.~C. Murphy, J.~L. Lund, E.~S. Dellon, J.~L. Williams, E.~T. Jensen, N.~J. Shaheen, A.~S. Barritt, S.~R. Lieber \emph{et~al.}, ``Burden and cost of gastrointestinal, liver, and pancreatic diseases in the united states: update 2018,'' \emph{Gastroenterology}, vol. 156, no.~1, pp. 254--272, 2019.

\bibitem{gunjan2014small}
D.~Gunjan, V.~Sharma, S.~S. Rana, and D.~K. Bhasin, ``Small bowel bleeding: a comprehensive review,'' \emph{Gastroenterology report}, vol.~2, no.~4, pp. 262--275, 2014.

\bibitem{nadler2014role}
M.~Nadler and R.~Eliakim, ``The role of capsule endoscopy in acute gastrointestinal bleeding,'' \emph{Therapeutic advances in gastroenterology}, vol.~7, no.~2, pp. 87--92, 2014.

\bibitem{kitiyakara2005non}
T.~Kitiyakara and W.~Selby, ``Non-small-bowel lesions detected by capsule endoscopy in patients with obscure gi bleeding,'' \emph{Gastrointestinal endoscopy}, vol.~62, no.~2, pp. 234--238, 2005.

\bibitem{thakur2024vce}
A.~Thakur, P.~Handa, N.~Goel, and D.~Gunjan, ``Vce-anomalynet: A new dataset fueling ai precision in anomaly detection for video capsule endoscopy⋆,'' \emph{Authorea Preprints}, 2024.

\bibitem{cortegoso2022inter}
P.~Cortegoso~Valdivia, U.~Deding, T.~Bj{\o}rsum-Meyer, G.~Baatrup, I.~Fern{\'a}ndez-Uri{\'e}n, X.~Dray, P.~Boal-Carvalho, P.~Ellul, E.~Toth, E.~Rondonotti \emph{et~al.}, ``Inter/intra-observer agreement in video-capsule endoscopy: are we getting it all wrong? a systematic review and meta-analysis,'' \emph{Diagnostics}, vol.~12, no.~10, p. 2400, 2022.

\bibitem{b3}
H.~S. Pannu, S.~Ahuja, N.~Dang, S.~Soni, and A.~K. Malhi, ``Deep learning based image classification for intestinal hemorrhage,'' \emph{Multimedia Tools and Applications}, vol.~79, pp. 21\,941--21\,966, 2020.

\bibitem{b2}
A.~Caroppo, A.~Leone, and P.~Siciliano, ``Deep transfer learning approaches for bleeding detection in endoscopy images,'' \emph{Computerized Medical Imaging and Graphics}, vol.~88, p. 101852, 2021.

\bibitem{charfi2024abnormalities}
S.~Charfi, M.~El~Ansari, L.~Koutti, A.~Ellahyani, and I.~Eljaafari, ``Abnormalities detection from wireless capsule endoscopy images based on embedding learning with triplet loss,'' \emph{Multimedia Tools and Applications}, pp. 1--22, 2024.

\bibitem{amiri2024combining}
Z.~Amiri, H.~Hassanpour, and A.~Beghdadi, ``Combining deep features and hand-crafted features for abnormality detection in wce images,'' \emph{Multimedia Tools and Applications}, vol.~83, no.~2, pp. 5837--5870, 2024.

\bibitem{lafraxo2024computer}
S.~Lafraxo, M.~El~Ansari, and L.~Koutti, ``Computer-aided system for bleeding detection in wce images based on cnn-gru network,'' \emph{Multimedia Tools and Applications}, vol.~83, no.~7, pp. 21\,081--21\,106, 2024.

\bibitem{kaur2023wireless}
P.~Kaur and R.~Kumar, ``Wireless capsule endoscopy video summarization using transfer learning and random forests,'' \emph{International Journal of Advanced Computer Science and Applications}, vol.~14, no.~9, 2023.

\bibitem{padmavathi2023wireless}
P.~Padmavathi and J.~Harikiran, ``Wireless capsule endoscopy infected images detection and classification using mobilenetv2-bilstm model,'' \emph{International Journal of Image and Graphics}, vol.~23, no.~05, p. 2350041, 2023.

\bibitem{sreejesh2023bleeding}
G.~Sreejesh, ``Bleeding frame and region detection in wireless capsule endoscopy video,'' \emph{International Journal of Human Computations \& Intelligence}, vol.~2, no.~1, pp. 26--33, 2023.

\bibitem{padmavathi2023effective}
P.~Padmavathi, J.~Harikiran, and J.~Vijaya, ``Effective deep learning based segmentation and classification in wireless capsule endoscopy images,'' \emph{Multimedia Tools and Applications}, pp. 1--25, 2023.

\bibitem{lafraxo2023semantic}
S.~Lafraxo, M.~Souaidi, M.~El~Ansari, and L.~Koutti, ``Semantic segmentation of digestive abnormalities from wce images by using attresu-net architecture,'' \emph{Life}, vol.~13, no.~3, p. 719, 2023.

\bibitem{bordbar2023wireless}
M.~Bordbar, M.~S. Helfroush, H.~Danyali, and F.~Ejtehadi, ``Wireless capsule endoscopy multiclass classification using 3d deep cnn model,'' \emph{Journal Name Not Available}, vol. N/A, no. N/A, p. N/A, 2023.

\bibitem{naz2023comparative}
J.~Naz, M.~I. Sharif, M.~I. Sharif, S.~Kadry, H.~T. Rauf, and A.~E. Ragab, ``A comparative analysis of optimization algorithms for gastrointestinal abnormalities recognition and classification based on ensemble xcepnet23 and resnet18 features,'' \emph{Biomedicines}, vol.~11, no.~6, p. 1723, 2023.

\bibitem{singh2022explainable}
A.~Singh, H.~S. Pannu, and A.~Malhi, ``Explainable information retrieval using deep learning for medical images,'' \emph{Computer Science and Information Systems}, vol.~19, no.~1, pp. 277--307, 2022.

\bibitem{mohankumar2022rnn}
C.~Mohankumar, S.~D. Kumar, S.~Senthilkumar, V.~Sornagopal, and M.~Maharajan, ``Rnn model-based classification of wireless capsule endoscopy bleeding images,'' \emph{International journal of health sciences}, no.~I, pp. 7330--7344, 2022.

\bibitem{deng2009imagenet}
J.~Deng, W.~Dong, R.~Socher, L.-J. Li, K.~Li, and L.~Fei-Fei, ``Imagenet: A large-scale hierarchical image database,'' in \emph{2009 IEEE conference on computer vision and pattern recognition}.\hskip 1em plus 0.5em minus 0.4em\relax Ieee, 2009, pp. 248--255.

\bibitem{redlesion}
{Red Lesion Endoscopy Dataset - CKAN}, ``Red lesion endoscopy dataset - ckan,'' \url{https://rdm.inesctec.pt/dataset/nis-2018-003}, May 2018, retrieved November 19, 2023.

\bibitem{deeba2016automated}
F.~Deeba, F.~M. Bui, and K.~A. Wahid, ``Automated growcut for segmentation of endoscopic images,'' in \emph{2016 International Joint Conference on Neural Networks (IJCNN)}.\hskip 1em plus 0.5em minus 0.4em\relax IEEE, 2016, pp. 4650--4657.

\bibitem{koulaouzidis2017kid}
A.~Koulaouzidis, D.~K. Iakovidis, D.~E. Yung, E.~Rondonotti, U.~Kopylov, J.~N. Plevris, E.~Toth, A.~Eliakim, G.~W. Johansson, W.~Marlicz \emph{et~al.}, ``Kid project: an internet-based digital video atlas of capsule endoscopy for research purposes,'' \emph{Endoscopy international open}, vol.~5, no.~06, pp. E477--E483, 2017.

\bibitem{smedsrud2021kvasir}
P.~H. Smedsrud, V.~Thambawita, S.~A. Hicks, H.~Gjestang, O.~O. Nedrejord, E.~N{\ae}ss, H.~Borgli, D.~Jha, T.~J.~D. Berstad, S.~L. Eskeland \emph{et~al.}, ``Kvasir-capsule, a video capsule endoscopy dataset,'' \emph{Scientific Data}, vol.~8, no.~1, p. 142, 2021.

\bibitem{coelho2018deep}
P.~Coelho, A.~Pereira, A.~Leite, M.~Salgado, and A.~Cunha, ``A deep learning approach for red lesions detection in video capsule endoscopies,'' in \emph{Image Analysis and Recognition: 15th International Conference, ICIAR 2018, P{\'o}voa de Varzim, Portugal, June 27--29, 2018, Proceedings 15}.\hskip 1em plus 0.5em minus 0.4em\relax Springer, 2018, pp. 553--561.

\bibitem{he2016identity}
K.~He, X.~Zhang, S.~Ren, and J.~Sun, ``Identity mappings in deep residual networks,'' in \emph{Computer Vision--ECCV 2016: 14th European Conference, Amsterdam, The Netherlands, October 11--14, 2016, Proceedings, Part IV 14}.\hskip 1em plus 0.5em minus 0.4em\relax Springer, 2016, pp. 630--645.

\bibitem{nolan2010inception}
C.~Nolan, \emph{Inception: The shooting script}.\hskip 1em plus 0.5em minus 0.4em\relax Insight Editions, 2010.

\bibitem{huang2017densely}
G.~Huang, Z.~Liu, L.~Van Der~Maaten, and K.~Q. Weinberger, ``Densely connected convolutional networks,'' in \emph{Proceedings of the IEEE conference on computer vision and pattern recognition}, 2017, pp. 4700--4708.

\bibitem{liu2022convnet}
Z.~Liu, H.~Mao, C.-Y. Wu, C.~Feichtenhofer, T.~Darrell, and S.~Xie, ``A convnet for the 2020s,'' in \emph{Proceedings of the IEEE/CVF conference on computer vision and pattern recognition}, 2022, pp. 11\,976--11\,986.

\bibitem{chollet2017xception}
F.~Chollet, ``Xception: Deep learning with depthwise separable convolutions,'' in \emph{Proceedings of the IEEE conference on computer vision and pattern recognition}, 2017, pp. 1251--1258.

\bibitem{simonyan2013deep}
K.~Simonyan, A.~Vedaldi, and A.~Zisserman, ``Deep inside convolutional networks: Visualising image classification models and saliency maps,'' \emph{arXiv preprint arXiv:1312.6034}, 2013.

\bibitem{zoph2018learning}
B.~Zoph, V.~Vasudevan, J.~Shlens, and Q.~V. Le, ``Learning transferable architectures for scalable image recognition,'' in \emph{Proceedings of the IEEE conference on computer vision and pattern recognition}, 2018, pp. 8697--8710.

\bibitem{howard2017mobilenets}
A.~Howard, ``Mobilenets: Efficient convolu-tional neural networks for mobile vision applications,'' \emph{arXiv preprint arXiv:1704.04861}, 2017.

\bibitem{simonyan2014very}
K.~Simonyan, ``Very deep convolutional networks for large-scale image recognition,'' \emph{arXiv preprint arXiv:1409.1556}, 2014.

\bibitem{redmon2016you}
J.~Redmon, S.~Divvala, R.~Girshick, and A.~Farhadi, ``You only look once: Unified, real-time object detection,'' in \emph{Proceedings of the IEEE conference on computer vision and pattern recognition}, 2016, pp. 779--788.

\bibitem{ronneberger2015u}
O.~Ronneberger, P.~Fischer, and T.~Brox, ``U-net: Convolutional networks for biomedical image segmentation,'' in \emph{Medical image computing and computer-assisted intervention--MICCAI 2015: 18th international conference, Munich, Germany, October 5-9, 2015, proceedings, part III 18}.\hskip 1em plus 0.5em minus 0.4em\relax Springer, 2015, pp. 234--241.

\bibitem{badrinarayanan2015segnet}
V.~Badrinarayanan, A.~Handa, and R.~Cipolla, ``Segnet: A deep convolutional encoder-decoder architecture for robust semantic pixel-wise labelling,'' \emph{arXiv preprint arXiv:1505.07293}, 2015.

\bibitem{chaurasia2017linknet}
A.~Chaurasia and E.~Culurciello, ``Linknet: Exploiting encoder representations for efficient semantic segmentation,'' in \emph{2017 IEEE visual communications and image processing (VCIP)}.\hskip 1em plus 0.5em minus 0.4em\relax IEEE, 2017, pp. 1--4.

\bibitem{kora2022transfer}
P.~Kora, C.~P. Ooi, O.~Faust, U.~Raghavendra, A.~Gudigar, W.~Y. Chan, K.~Meenakshi, K.~Swaraja, P.~Plawiak, and U.~R. Acharya, ``Transfer learning techniques for medical image analysis: A review,'' \emph{Biocybernetics and Biomedical Engineering}, vol.~42, no.~1, pp. 79--107, 2022.

\bibitem{alzubaidi2021novel}
L.~Alzubaidi, M.~Al-Amidie, A.~Al-Asadi, A.~J. Humaidi, O.~Al-Shamma, M.~A. Fadhel, J.~Zhang, J.~Santamar{\'\i}a, and Y.~Duan, ``Novel transfer learning approach for medical imaging with limited labeled data,'' \emph{Cancers}, vol.~13, no.~7, p. 1590, 2021.

\bibitem{qureshi2023comprehensive}
R.~Qureshi, M.~G. Ragab, S.~J. Abdulkader, A.~Alqushaib, E.~H. Sumiea, H.~Alhussian \emph{et~al.}, ``A comprehensive systematic review of yolo for medical object detection (2018 to 2023),'' \emph{Authorea Preprints}, 2023.

\bibitem{yin2022u}
X.-X. Yin, L.~Sun, Y.~Fu, R.~Lu, and Y.~Zhang, ``U-net-based medical image segmentation,'' \emph{Journal of healthcare engineering}, vol. 2022, 2022.

\bibitem{goel2022dilated}
N.~Goel, S.~Kaur, D.~Gunjan, and S.~Mahapatra, ``Dilated cnn for abnormality detection in wireless capsule endoscopy images,'' \emph{Soft Computing}, pp. 1--17, 2022.

\bibitem{liu2009obscure}
J.~Liu and X.~Yuan, ``Obscure bleeding detection in endoscopy images using support vector machines,'' \emph{Optimization and engineering}, vol.~10, no.~2, pp. 289--299, 2009.

\bibitem{o2023capsule}
F.~J. O'Hara and D.~Mc~Namara, ``Capsule endoscopy with artificial intelligence-assisted technology: Real-world usage of a validated ai model for capsule image review,'' \emph{Endoscopy International Open}, vol.~11, no.~10, pp. E970--E975, 2023.

\bibitem{haslach2023reading}
M.~Haslach-H{\"a}fner and K.~M{\"o}nkem{\"u}ller, ``Reading capsule endoscopy: Why not ai alone?'' \emph{Endoscopy International Open}, vol.~11, no.~12, pp. E1175--E1176, 2023.

\end{thebibliography}

\end{document}